%% file: main.tex
\documentclass[10pt,twocolumn,letterpaper]{article}

\usepackage{cvpr}              

\usepackage[accsupp]{axessibility}  


\definecolor{cvprblue}{rgb}{0.21,0.49,0.74}
\usepackage[pagebackref,breaklinks,colorlinks,allcolors=cvprblue]{hyperref}

\usepackage{algorithm}
\usepackage{booktabs}
\usepackage{multirow}
\usepackage{array}
\usepackage{pifont}
\usepackage{rotating}
\usepackage{graphicx, subcaption, overpic, textpos}
\newlength\savewidth

\newcolumntype{x}[1]{>{\centering\arraybackslash}p{#1pt}}

\usepackage{algpseudocode} 
\usepackage{amsmath}
\usepackage{amssymb}
\usepackage{mathrsfs}

\newtheorem{theorem}{Theorem}

\newtheorem{definition}{Definition}

\def\paperID{10749} 
\def\confName{CVPR}
\def\confYear{2025}

\title{GeoMM: On Geodesic Perspective for Multi-modal Learning}

\author{
	Shibin Mei\textsuperscript{\rm 1} \quad \quad 
	Hang Wang\textsuperscript{\rm 1} \quad \quad 
	Bingbing Ni\textsuperscript{\rm 1,2}\thanks{Corresponding author: Bingbing Ni.} 
        \\
	\textsuperscript{\rm 1}Huawei   \quad
	\textsuperscript{\rm 2}Shanghai Jiao Tong University 
   \\ \{shibin.mei1027, \, francis970625\}@gmail.com \,
	nibingbing@sjtu.edu.cn
}

\begin{document}
\maketitle

\begin{abstract}
Geodesic distance serves as a reliable means of measuring distance in nonlinear spaces, and such nonlinear manifolds are prevalent in the current multimodal learning. In these scenarios, some samples may exhibit high similarity, yet they convey different semantics, making traditional distance metrics inadequate for distinguishing between positive and negative samples. This paper introduces geodesic distance as a novel distance metric in multi-modal learning for the first time, to mine correlations between samples, aiming to address the limitations of common distance metric. Our approach incorporates a comprehensive series of strategies to adapt geodesic distance for the current multimodal learning. Specifically, we construct a graph structure to represent the adjacency relationships among samples by thresholding distances between them and then apply the shortest-path algorithm to obtain geodesic distance within this graph. To facilitate efficient computation, we further propose a hierarchical graph structure through clustering and combined with incremental update strategies for dynamic status updates. Extensive experiments across various downstream tasks validate the effectiveness of our proposed method, demonstrating its capability to capture complex relationships between samples and improve the performance of multimodal learning models.

\end{abstract}


\section{1 Introduction}
\label{sec_intro}

Recent years have witnessed remarkable progress on multimodal learning~\cite{li2019visualbert, lu2019vilbert, radford2021learning, li2021align, bao2022vlmo}, driven by the rapid development of large transformer models~\cite{vaswani2017attention}. In multimodal learning, image-text pairs are encoded into a unified representation space through carefully designed pre-training tasks~\cite{chen2020uniter, li2020oscar, zhang2021vinvl}. The consistency of cross-modal representations facilitates the model with strong generalization abilities~\cite{yu2022coca, wang2022image}, achieving satisfactory performance on various downstream tasks.

\begin{figure}[t]  
    \centering  
\includegraphics[width=0.5\textwidth]{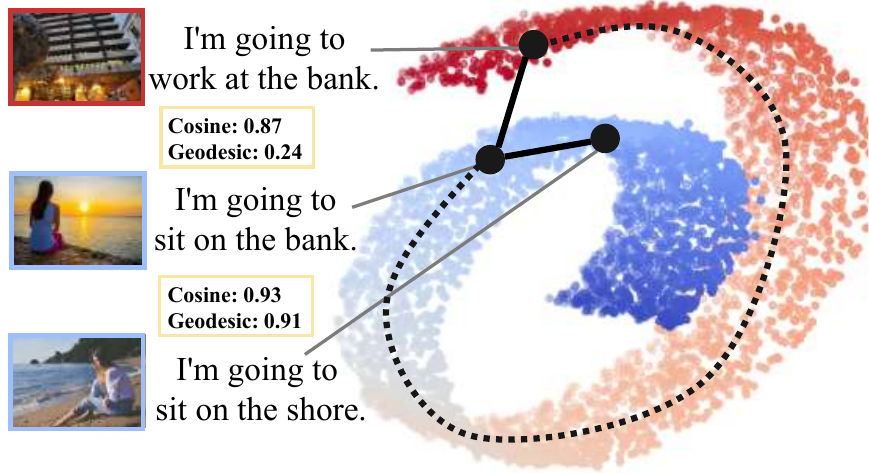}   
    \caption{The motivation of leveraging geodesic metric. Based on the topological structures, geodesic distance can depict sample similarity more accurately.}  
    \label{fig_motivation}  
\end{figure}

A prevalent paradigm in multimodal learning includes extracting features from different modalities, encouraging proximity between matching features and dispersity between mismatched ones by contrastive supervision~\cite{radford2021learning}, such as InfoNCE loss~\cite{he2020momentum}. This paradigm inherently relies on distance computation between multi-modality features to mine positive and negative samples. CLIP~\cite{radford2021learning} computes cosine similarity between cross-modal samples within a large batch, while ALBEF~\cite{li2021align}, maintains a momentum feature queue and calculates distances between batch features and queue features, thus avoiding large batch size. Recent works, such as TCL~\cite{yang2022vision} and MAFA~\cite{byun2024mafa}, adopt a similar paradigm but pay more attention to negatives. However, these works assume that samples are distributed in a spherical space~\cite{he2020momentum}, overlooking the possibility of distributions in more complex non-Euclidean spaces. For example, some sentences similar at the word level may possess close distances with trivial metrics but convey vastly different semantics, as shown in Fig.\ref{fig_motivation}. We propose that geodesic distance~\cite{kimmel1998computing} can serve as an effective distance metric for multimodal learning, aiding in considering the data topology structure and thus more accurately depicting distances between samples.

In this paper, we introduce geodesic distance to multimodal learning for the first time. 
Such distance metric enhances the characterization of relationships between samples in multi-modality datasets with complex data distribution. The benefits of using geodesic distance as a metric in multimodal learning are evident. Firstly, geodesic distance more accurately captures the underlying structure of complex data manifolds, especially when integrating data from diverse modalities.
Therefore, we can easily distinguish samples with close cosine distances but distant geodesic distances to mine more accurate positive and negative samples. 
Secondly, traditional distance computations are one-to-one, meaning the distance between two points depends solely on those two points, which can lead to inaccuracy in portraying relative distances between distant points~\cite{chowdhury2022unsupervised}.
This ambiguity can adversely impact model optimization. In contrast, geodesic distance considers many-to-many (global) relationships, where data points are interconnected and the distance between two samples involves the participation of other routing points. The deterministic point-to-point path significantly reduces ambiguity and enhances interpretability in distance calculations, especially for more distant samples.

Unlike traditional distance, the calculation of geodesic distance relies on a sample pool, which provides a reference frame for modeling distances between sample pairs. We posit that the local spaces neighboring each sample adhere to simple manifold assumption (as Def.\ref{def_simple_manifold}), allowing us to compute distances using a straightforward one-to-one metric (as Def.\ref{def_trivial_distance}). 
Consequently, we build a graph based on short-distance adjacency relationships within the sample pool, establishing undirected edges between locally adjacent samples. The distance between samples is represented as the shortest path between points in this graph, which can be computed using shortest-path algorithms. For simplicity, we employ the Floyd algorithm~\cite{floyd1962algorithm}.
However, directly calculating geodesic distance for a large number of samples would result in huge computational complexity. To facilitate efficient computation, we model the sample pool as a hierarchical graph. Within each layer of the graph, we partition samples into disjoint clusters through sample clustering. We then construct a graph among cluster centers and use the Floyd algorithm to calculate the shortest path between these cluster centers. Distances within each cluster are recursively propagated to the next layer of the hierarchical graph. At the bottom layer, we assume that the local manifold is simple enough to employ trivial distance metrics. For the distance between any two samples, we backtrack layer by layer in the hierarchical graph from bottom to top, to assess whether they belong to the same cluster, up to the top layer, as illustrated in Fig.\ref{fig_overview}. Additionally, we develop efficient incremental update strategies to accommodate sample updates and reduce computational complexity. 

We believe geodesic distance can effectively facilitate current multimodal learning, yielding improved pre-trained models with more accurate distance modeling.
Our multimodal pre-training experiments yield promising results across various tasks, such as image-text retrieval, visual question answering, visual entailment, natural language for visual reasoning, and visual grounding. Extensive results verify the effectiveness of this geodesic perspective.

\section{Related Work}

\subsection{Multi-modal Learning}
Multimodal pre-training can be mainly categorized into single-stream networks~\cite{li2019visualbert, lu2019vilbert, su2019vl, chen2020uniter, li2020oscar, pixelbert}, dual-stream networks~\cite{TanB19,radford2021learning,li2021align,jia2021scaling,li2023scaling}, and Xformer networks~\cite{yu2022coca,li2022blip,bao2022vlmo,mei2025object,chen2021x}. Single-stream networks concatenate text tokens and visual tokens and then utilize a transformer~\cite{vaswani2017attention} for modality fusion. 
This structure can be seen as treating visual modality as prompt for the text modality~\cite{li2023multimodal}. 
Depending on the way of obtaining visual tokens, single-stream networks can further be divided into region-based methods~\cite{chen2020uniter, li2020oscar, zhang2021vinvl}, grid-based methods~\cite{pixelbert}, and patch-based methods~\cite{kim2021vilt}. 
Dual-stream networks encode images and texts separately and then align paired image-text features with contrastive loss to map features of different modalities into a unified representation space. This paradigm can be viewed as maximizing a lower bound on the mutual information between modalities~\cite{li2021align}. 
Among them, CLIP~\cite{radford2021learning}, trained on 400M image-text pairs, has demonstrated remarkable zero-shot ability across various downstream tasks.
Subsequent improvement efforts include~\cite{jia2021scaling,pham2023combined, li2023scaling, chen2023stair}. ImageBind~\cite{girdhar2023imagebind} aligns more modalities and has inspired numerous interesting works~\cite{su2023pandagpt}. Xformer networks combine advantages from single-stream and dual-stream networks, performing end-to-end feature extraction, alignment, and fusion for different modalities. 
Benefiting from the design, Xformer exhibits strong retrieval capabilities as well as superior multimodal reasoning ability. 
Numerous works have also introduced novel structures such as VLMO~\cite{bao2022vlmo} and BeiT3~\cite{wang2022image}. 
Multimodal learning requires elaborate pre-training tasks, and commonly used pre-training tasks include Masked Language Modeling(MLM)~\cite{devlin2018bert}, Image-Text Matching (ITM)~\cite{li2021align}, Image-Text Contrastive (ITC)~\cite{radford2021learning}, Word Patch Alignment (WPA)~\cite{chen2020uniter}, and so on.

\subsection{Geodesic Distance}

Geodesic distance provides more structural information in complex manifolds. Mitchell et al.~\cite{MitchellMP87} proposes an MMP algorithm to get the exact computation of geodesic distance. Many approximate methods~\cite{kimmel1998computing, SurazhskySKGH05,CraneWW13} are developed to reduce computational cost.
The classic dimensionality reduction algorithm ISOMAP~\cite{isomap} uses geodesic distance to measure the distance between points, and then reconstructs the data in a low-dimensional Euclidean space. 
Recently, many arts of various fields have integrated geodesic distance into existing deep learning paradigms and achieved favorable results. GCNN~\cite{MasciBBV15} proposes a convolution operator that adapts to the data manifold, taking into account the manifold distribution within the patch during the feature extraction. To reduce computational complexity, GeoNet~\cite{HeHYZWWS19} is committed to learning potential sample geodesic features. 
Geodesic-Former~\cite{NgoN22} proposes a geometric-guided approach to enhance the few-shot capability in point cloud segmentation.
GraphWalk~\cite{PotamiasNBZ22} proposes a differentiable geodesic distance estimator, but suffers from low accuracy. Except for the application in 3D vision~\cite{NgoN22,LiTXWYCW22}, there are also explorations on image~\cite{BaiS09,LingJ05,LeNYS16} to achieve appearance-based similarity estimates. 
In this paper, we use geodesic distance for contrastive supervision for the first time in multimodal learning seamlessly and calculate geodesic distance based on topological structures.

\section{Methodology}

In this section, we elaborate on employing geodesic distance as distance measurement, and utilizing multimodal learning as a point of entry. Sec.3.1 introduces basic paradigms of distance computation in multimodal learning. In Sec.3.2 delves into the algorithms of geodesic distance computation. Next, in Sec.3.3 and Sec.3.4, we develop efficient methods for geodesic metric computation and the adaption to current multimodal learning. Finally, Sec. 3.5 and Sec. 3.6 detail the algorithmic implementation and theoretical analyses.

\begin{figure}[t]  
    \centering  
\includegraphics[width=0.46\textwidth]{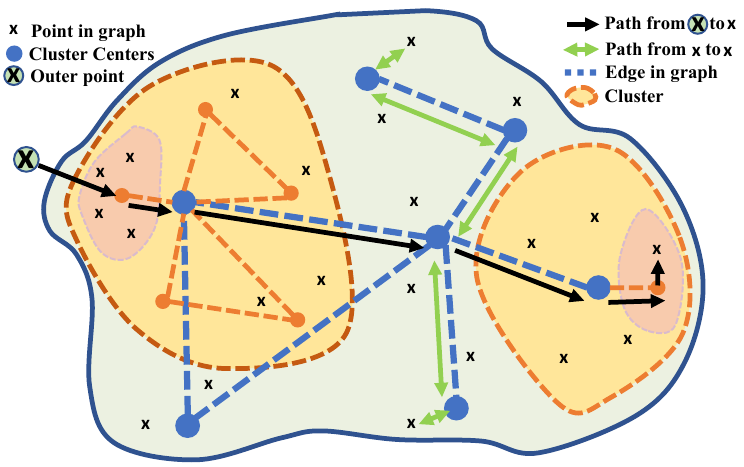}  
    \caption{An overview of our proposed hierarchical graph structure, where we display the clusters and geodesic paths.}  
    \label{fig_overview}  
\end{figure}

\subsection{Preliminary}
\label{sec_overview}

In multimodal learning, a fundamental paradigm involves extracting features from numerous number of paired samples (e.g. image-text pairs), then pulling paired samples closer while pushing mismatched samples apart~\cite{radford2021learning, jia2021scaling,mei2022towards}. Stable model optimization relies on a sufficient number of negative samples for contrastive learning~\cite{chen2020simple, he2020momentum,wang2020learning,mei2023exploring,mei2024mssidd,mei2023towards}. For instance, CLIP~\cite{radford2021learning} employs a very large batch size of 32,768, while ALBEF~\cite{li2021align}, TCL~\cite{yang2022vision} and MAFA~\cite{byun2024mafa} maintains a large and consistent momentum feature queue with size of 65,536, 65,536 and 480,000, respectively. All these methods provide sufficient samples for constructing topological structures, which enables precise computation of geodesic distances. Considering computational efficiency and model performance, we adopt the paradigm of ALBEF, TCL and MAFA. The feature queue generated by the momentum model ensures an adequate number of contrastive samples even with small batch sizes. As presented in Alg.\ref{alg_single_step}, 
given the momentum feature queue $Q_a, Q_b\in\mathbb{R}^{N\times c}$, batch feature $F_a, F_b\in\mathbb{R}^{B\times c}$ of two modalities, where $N$ is the feature queue size, $B$ denotes batch size, $c$ is channel dimension, the distance between $F_a$ and $Q_b$ necessitates computing the geodesic distances from each sample in $F_a$ to each sample in the feature queue $Q_b$, and vice versa.
Additionally, we need to update the current batch into the feature queue without disrupting the topological structure used for computing geodesic distances.

\subsection{Geodesic View}
\label{sec_geodesic}

The computation of geodesic distance typically relies on assumptions about the manifold structure. A common assumption~\cite{andruchow1999geometry} is that data is intrinsically embedded onto a spherical surface~\cite{he2020momentum}. 
However, this assumption could be overly simplistic, as mapping samples onto a spherical surface fails to capture more complex manifold structures, as shown in Fig.\ref{fig_overview}.
To better accommodate the complex manifold, we opt for data topological structures for geodesic distance computation.
Previous works~\cite{isomap,NgoN22} construct feature points into a graph using local proximity relationships. Inspired by this, for a sample pool $P$ of size $N$, we define each point to possess proximity relationships with its $n$ nearest neighbors.

\begin{algorithm}[!t]  
\caption{A Step for Multimodal Model Optimization}  
\label{alg_single_step}  
\begin{algorithmic}[1]  
		\Require
		Data of modalities $A$, $B$, feature queue $Q_a$, $Q_b$
		\Ensure 
		Model parameters $\theta$
		\State Extract features $F_a$, $F_b$ and moment features $F^m_a$, $F^m_b$ for input batch.
		\State Compute the distance between $F_a$ and $[F^m_b, Q_b]$, and the distance between $F_b$ and $[F^m_a, Q_a]$.
		\State Calculate losses and update model parameters $\theta$.
		\State Update $F^m_a$ and $F^m_b$ into feature queues $Q_a$ and $Q_b$.	
	\end{algorithmic}
\end{algorithm}

For the sake of discussion, we first give some basic definitions as follows,
\begin{definition}
\label{def_trivial_distance}
(Trivial Distance Metric) We define the distance metric as the trivial distance metric if and only if the distance calculation relies solely on these two related points, and is independent of other routing points, such as cosine distance and Euclidean distance.
\end{definition}

\begin{definition}
\label{def_simple_manifold}
(Simple Manifold Assumption) For any two points $x_i, x_j$ within a local space with flat data distribution, the difference between geodesic metric $d_g(x_i, x_j)$ and trivial metric $d_t(x_i, x_j)$ satisfies,
\begin{equation}
\lvert d_g(x_i, x_j) - d_t(x_i, x_j) \rvert < \delta,
\end{equation}
where $\delta$ is a threshold. We can then define this local space satisfies simple manifold assumption. 
\end{definition}
For two proximity points $x_i$ and $x_j$, there exists an edge with a weight of $d(x_i, x_j)$ between them. We assume the local manifold surrounding each point to be a simple manifold, where we can employ a trivial distance metric. Consequently, all points are encompassed within a graph structure, and we can utilize the Floyd algorithm~\cite{floyd1962algorithm} to compute the shortest paths between each point pair. Ultimately, we obtain a geodesic distance matrix $G_D\in\mathbb{R}^{N\times N}$ with respect to the sample pool $P$.

\subsection{Hierarchical Structure}
\label{sec_hierarchical}

Despite the above geodesic perspectives, applying geodesic distance in multimodal learning remains several challenges. Firstly, to ensure stable model convergence, feature queues are often large, resulting in a considerably large sample pool. Constructing a graph and computing the shortest paths for all points in such a pool can be computationally prohibitive. Secondly, the distances from points outside the graph to every point within the graph must also be taken into account. Finally, the momentum features of each batch should be updated into the graph after each step optimization. We will elaborate on the solutions to these issues in detail below.

\textbf{\textit{Q1: How to obtain geodesic distances between points in a large-scale sample pool?}}

Based on the simple manifold assumption of local data space, we first cluster the samples in the sample pool, with the number of cluster centers denoted as $n_{cluster}$. Subsequently, we only construct graphs for the cluster centers and compute the shortest paths by the Floyd algorithm. At this point, the distance between any two points $x_i, x_j$ in the sample pool can be expressed as the distance from each point to its respective cluster center $C(x_i), C(x_j)$, denoted as $d(x_i, C(x_i)), d(x_j, C(x_j))$, plus the shortest path distance between the two cluster centers, denotes as $d_g(C(x_i), C(x_j))$. The distance computation from a sample point to its corresponding cluster center occurs within a cluster. If the cluster is small enough to meet the simple manifold assumption, we can directly apply a trivial distance metric. Otherwise, we recursively apply clustering within this cluster until the clusters at the lowest level satisfy the simple manifold assumption. 
Consequently, the distances between any two points in the sample pool proceed by traversing the hierarchical structure of the graph from bottom to top, accumulating the distances between sample points and their respective cluster centers during the backtracking process, until the two cluster centers at a certain level belong to the same cluster. Then, the geodesic distance between these two cluster centers will be added to the final result, as shown in Fig.~\ref{fig_overview}. We name this algorithm as \textbf{hierarchical geodesic distance algorithm}. 
For a graph with $n$ layers and the $n$-th layer is the bottommost layer, the distance of $d(x_i, x_j)$ can be calculated as
\begin{equation}
\small
\begin{split}
\label{eq_dxixj}
d(x_i, x_j)=& d_g(C_s(x_i), C_s(x_j)) + d(x_i, C_s(x_i)) + d(x_j, C_s(x_j)),
\\  \text{where}\quad\quad\;\;& \\
 d(x_i,  C_k(x_i&))=d_g(C_{k+1}(x_i), C_{k+1}(C_k(x_i))) \\ & + d(x_i, C_k(x_i)) \\ & + d(C_k(x_i), C_{k+1}(C_k(x_i))),  \text{when} \; s \leq k < n \\
 d(x_i, C_n(x_i&))=d_t(x_i, C_n(x_i)), \;\text{when}\; k=n.
\end{split}
\end{equation}
$s$ denotes the layer where two sub-cluster centers belong to the same cluster. $C_s, C_{s+1}, ..., C_n$ denote the corresponding cluster centers at layer $s,s+1,...,n$. $d_g()$ represents geodesic distance obtained by Floyd algorithm, while $d_t$ represents trivial distance metrics, such as cosine distance and Euclidean distance.

\textbf{\textit{Q2: How to compute the distance from a point outside the graph to all points in the sample pool?}}

For the constructed hierarchical graph, the distance from each point $x_i$ to its respective bottom-level cluster center $C_n(x_i)$, and the distance among all bottom-level cluster centers in the graph, can be conveniently obtained. By combining these two distances, we obtain the distance from each point $x_i$ in the graph to all bottom-level cluster centers $\mathcal{C}_n$, i.e. $d(x_i,C_n(x_j))=d(x_i,C_n(x_i))+d(C_n(x_i),C_n(x_j))$, where $C_n(x_j)$ is any one bottom level cluster center. Subsequently, we decompose the distance from a point outside the graph $x_o$ to a point in the sample pool $x_i$ into the distance from the point outside the graph to its nearest bottom-level cluster center $d(x_o, C_n^{near}(x_o))$ and the distance from this cluster center to the point inside the graph $d(C_n^{near}(x_o), x_i)$, as shown in Fig.\ref{fig_overview}. We have already obtained the distance from cluster centers to any points inside the graph $d(C_n, x_i)$ as described above. Hence, we ultimately obtain the distance from a point outside the graph to any point inside the graph, as Eq.\ref{eq_out_to_in},
\begin{equation}
\label{eq_out_to_in}
d(x_o, x_i) = d(x_o, C_n^{near}(x_o)) + d(C_n^{near}(x_o), x_i).
\end{equation}

\textbf{\textit{Q3: How to perform graph updates?}}

With our proposed hierarchical structure, graph updates become relatively straightforward. As mentioned above, we have already determined the nearest bottom-level cluster center for each point outside the graph. We simply need to attach these points to their respective nearest cluster centers. To ensure that the overall size of the graph remains stable, we remove the oldest samples following~\cite{he2020momentum}.

In addition to the feature queue, we maintain a circular index queue representing the bottom-level cluster centers to which each point belongs. The graph update boils down to covering the feature queue and the index queue. To avoid redundant computations, we also maintain queues for the distances from each point to its nearest bottom-level cluster center and to all bottom-level cluster centers, and we update these two queues simultaneously. To prevent the hierarchical structure from becoming outdated during training, we perform an overall update to the hierarchical graph structure every $T_0$ step. The overall training process can be summarized as follows. Firstly, in every $T_0$ step, we reconstruct the hierarchical graph structure $\mathcal{S}$, obtaining the bottom-level cluster center $\mathcal{K}$ for each point in the graph and the distances $\mathcal{D}_o$ from each point to all bottom-level centers. For a common step $t$, we fix the hierarchical graph $\mathcal{S}$, and compute the geodesic distances from each point in batch to the points in the sample pool for the model optimization. Finally, we update the $\mathcal{K}$ and $\mathcal{D}_o$ with the current batch.

\begin{table*}[t!]
\centering
\setlength\tabcolsep{4.5pt}
\begin{tabular}{l|c|cccccc|cccccc}
\toprule
\multicolumn{1}{c|}{\multirow{3}{*}{Method}} & \multirow{3}{*}{\#Images} & \multicolumn{6}{c|}{MSCOCO (5K test set)}                                                                                     & \multicolumn{6}{c}{Flickr30K  (1K test set)}                                                                \\
\multicolumn{1}{c|}{}                        &                           & \multicolumn{3}{c}{Text Retrieval}            & \multicolumn{3}{c|}{Image Retrieval}                                & \multicolumn{3}{c}{Text Retrieval}            & \multicolumn{3}{c}{Image Retrieval}           \\
\multicolumn{1}{c|}{}                        &                           & R@1           & R@5           & R@10          & R@1           & R@5           & \multicolumn{1}{c|}{R@10}          & R@1           & R@5           & R@10          & R@1           & R@5           & R@10          \\ \hline
ImageBERT~\cite{ImageBERT}                                      & 6M                        & 44.0          & 71.2          & 80.4          & 32.3          & 59.0          & 70.2          & 70.7          & 90.2          & 94.0          & 54.3          & 79.6          & 87.5         \\
UNITER~\cite{chen2020uniter}                                       & 4M                        & 64.1          & 87.7          & 93.3          & 48.8          & 76.7          & \multicolumn{1}{c|}{85.8}          & 80.7          & 95.7          & 98.0          & 66.2          & 88.4          & 92.9          \\
ViLT~\cite{kim2021vilt}                                      & 4M                        & 56.5          & 82.6          & 89.6          & 40.4          & 70.0          & \multicolumn{1}{c|}{81.1}          & 73.2          & 93.6          & 96.5          & 55.0          & 82.5          & 89.8          \\
CLIP~\cite{radford2021learning}                                       & 400M                      & 58.4          & 81.5          & 88.1          & 37.8          & 62.4          & \multicolumn{1}{c|}{72.2}          & 88.0          & 98.7          & 99.4          & 68.7          & 90.6          & 95.2          \\
FILIP~\cite{yao2021filip}                                         & 400M                      & 61.3          & 84.3          & 90.4          & 45.9          & 70.6          & 79.3          & 89.8          & 99.2          & \textbf{99.8}          & 75.0          & 93.4          & 96.3          \\
ALBEF~\cite{li2021align}                       & 4M                        & 68.7          & 89.5          & 94.7          & 50.1          & 76.4          & 84.5         & 90.5          & 98.8          & 99.7          & 76.8          & 93.7          & 96.7          \\
TCL~\cite{yang2022vision}                                       & 4M                        & 71.4          & 90.8          & 95.4          & 53.5          & 79.0          & 87.1          & 93.0          & 99.1          & 99.6          & 79.6          & 95.1          & 97.4          \\ 
MAFA*~\cite{byun2024mafa}                                         & 4M                        & 72.6          & 91.3          & 95.6         & 53.9          & 79.6          & 87.7          & 93.5          & 99.2          & 99.7          & 80.1         & 95.6          & 97.7          \\  \hline
\textbf{Geo-ALBEF}                                       & 4M                        & 72.0 & 91.1 & 95.6 & 53.6 & 79.2 & 87.2 & 93.2 & 99.3 & \textbf{99.8} & 79.9 & 95.3 & 97.5 \\
\textbf{Geo-TCL}                                       & 4M                        & 73.9 & 92.0 & 95.8 & 54.6 & 80.6 & 88.0 & 94.0 & \textbf{99.4} & \textbf{99.8} & 80.6 & 95.9 & 97.9 \\
\textbf{Geo-MAFA}                                      & 4M                        & \textbf{74.7} & \textbf{92.5} & \textbf{96.1} & \textbf{55.4} & \textbf{81.4} & \textbf{88.7} & \textbf{94.6} & \textbf{99.4} & \textbf{99.8} & \textbf{81.1} & \textbf{96.3} & \textbf{98.2} \\ \hline
ALIGN~\cite{jia2021scaling}                                      & 1.2B                      & 58.6          & 83.0          & 89.7          & 45.6          & 69.8          & \multicolumn{1}{c|}{78.6}          & 88.6          & 98.7          & 99.7          & 75.7          & 93.8          & 96.8          \\ 
\bottomrule
\end{tabular}
\caption{Zero-shot image-text retrieval results on MSCOCO and Flickr30K dataset. We report the average of R@1, R@5, R@10 for evaluation.  * refers to the reproduced models by the authors.}
\label{tab: zero_re}
\end{table*}


\begin{table*}[t!]
\centering
\setlength\tabcolsep{4.5pt}
\begin{tabular}{l|c|cccccc|cccccc}
\toprule
\multicolumn{1}{c|}{\multirow{3}{*}{Method}} & \multirow{3}{*}{\#Images} & \multicolumn{6}{c|}{MSCOCO (5K test set)}                                                                                     & \multicolumn{6}{c}{Flickr30K  (1K test set)}                                                                 \\
\multicolumn{1}{c|}{}                        &                           & \multicolumn{3}{c}{Text Retrieval}            & \multicolumn{3}{c|}{Image Retrieval}                                & \multicolumn{3}{c}{Text Retrieval}             & \multicolumn{3}{c}{Image Retrieval}           \\
\multicolumn{1}{c|}{}                        &                           & R@1           & R@5           & R@10          & R@1           & R@5           & \multicolumn{1}{c|}{R@10}          & R@1           & R@5           & R@10           & R@1           & R@5           & R@10          \\ \hline
ImageBERT~\cite{ImageBERT}                                     & 6M                        & 66.4          & 89.8          & 94.4          & 50.5          & 78.7          & 87.1          & 97.0          & 97.6          & 99.2          & 73.1          & 94.1          & 96.8         \\
UNITER~\cite{chen2020uniter}                                     & 4M                        & 65.7          & 88.6          & 93.8          & 52.9          & 79.9          & \multicolumn{1}{c|}{88.0}          & 87.3          & 98.0          & 99.2           & 75.6          & 94.1          & 96.8          \\
OSCAR~\cite{li2020oscar}                                       & 4M                        & 70.0          & 91.1          & 95.5          & 54.0          & 80.8          & \multicolumn{1}{c|}{88.5}          & -             & -             & -              & -             & -             & -             \\
ViLT~\cite{kim2021vilt}                                     & 4M                        & 61.5          & 86.3          & 92.7          & 42.7          & 72.9          & \multicolumn{1}{c|}{83.1}          & 83.5          & 96.7          & 98.6           & 64.4          & 88.7          & 93.8          \\
VILLA~\cite{gan2020large}                                      & 4M                        & -             & -             & -             & -             & -             & \multicolumn{1}{c|}{-}             & 87.9          & 97.5          & 98.8           & 76.3          & 94.2          & 96.8 \\
VLMO-B~\cite{bao2022vlmo}                                      & 4M                        & 74.8          & 93.1          & 96.9          & 57.2          & 82.6          & \multicolumn{1}{c|}{89.8}          & 92.3          & 99.4          & 99.9           & 79.3          & 95.7          & 97.8          \\
ALBEF~\cite{li2021align}                                       & 4M                        & 73.1          & 91.4          & 96.0          & 56.8          & 81.5          & \multicolumn{1}{c|}{89.2}          & 94.3          & 99.4          & 99.8           & 82.8          & 96.7          & 98.4          \\
TCL~\cite{yang2022vision}                                        & 4M                        & 75.6          & 92.8          & 96.7          & 59.0          & 83.2         & \multicolumn{1}{c|}{89.9}          & 94.9          & 99.5          & 99.8           &  84.0          & 96.7          & 98.5          \\
MAFA~\cite{byun2024mafa}                                       & 4M                        & 78.0          &   93.4       &    96.9     & 61.2          &  83.9    &   90.2      & 96.1         &      99.7    &     99.8     & 84.9         &   97.4     &  98.6        \\
\hline
\textbf{Geo-ALBEF}                                        & 4M       & 76.2 & 93.3 & 96.6 & 59.2 & 83.5 & 89.9 & 95.1 & 99.8 & 99.9 & 84.2 & 97.1 & 98.5 \\
\textbf{Geo-TCL}                                        & 4M       & 78.4 & 93.7 & 96.9 & 60.9 & 84.0 & 90.5 & 96.3 & \textbf{99.9} & 99.9 & 85.1 & 97.6 & 98.6 \\
\textbf{Geo-MAFA}                                     & 4M          & \textbf{79.3} & \textbf{93.9} & \textbf{97.3} & \textbf{62.5} & \textbf{84.3} & \textbf{90.8} & \textbf{96.9} & \textbf{99.9} & \textbf{100.0} & \textbf{85.6} & \textbf{98.1} & \textbf{98.9} \\ \hline
ALIGN~\cite{jia2021scaling}                                          & 1.2B                      & 77.0          & 93.5          & 96.9          & 59.9          & 83.3          & \multicolumn{1}{c|}{89.8}          & 95.3          & 99.8          & 100.0          & 84.9          & 97.4          & 98.6          \\ 
\bottomrule
\end{tabular}
\caption{Fine-tuned image-text retrieval results on MSCOCO and Flickr30K dataset.}
\label{tab: ft_re}
\end{table*}

\subsection{Algorithm Adaptation}
\label{sec_adapt}

In multimodal learning, we need to compute the distance from the features $F_a$ of modality $A$ to the features $F^m_b$ of modality $B$ as well as the feature queue $Q_b$. The paired samples from $F_a$ and $F^m_b$ serve as positive samples, while the feature queue $Q_b$ serves as negative samples. This requires us to first update the feature $F^m_b$ to $Q_b$, which involves treating $F^m_b$ as points outside the graph, computing the distance from each sample in $F^m_b$ to the sample pool $Q_b$, and then updating the intermediate results to the corresponding queue. Next, the features of modality $A$ are treated as points outside the graph, and the distances from each sample in $F_a$ to each point in the updated sample pool $Q_b^{update}$ are computed. Since the paired features in the sample pool and modality $A$ depend on the position where $F^m_b$ is inserted into the pool, we hence construct the corresponding ground truth based on the insertion position of $F^m_b$.

Under the assumption of a simple manifold, we use cosine distance as the trivial metric. Then each segment of the geodesic distance path corresponds to an angular distance, with the total geodesic distance representing the accumulated angles along the shortest path. We truncate the final accumulated value, normalize it to the range [0, $\pi$], and then apply the cosine function to compute the loss. We refer to this procedure as angle normalization. The complete algorithm is detailed in Alg.\ref{alg_geodesic_main}.

\begin{algorithm}[!t]  
	\caption{Hierarchical Geodesic Distance Adaptation for Multimodal Learning}  
	\label{alg_geodesic_main}  
	\begin{algorithmic}[1]  
        \small
		\Require
		Multimodal data of modalities $A$, $B$, feature queue $Q_a$, $Q_b$, total training steps $T$, hierarchical graph update period $T_0$
		\Ensure 
		Model parameters $\theta$
		\For{t=1 to $T$}
			\If{$t\; \% \; T_0$ == 0}
        	    \State Update the hierarchical graph $\mathcal{S}$.
				\State Obtain the bottom-layer clusters $\mathcal{K}$ for each point.
                    \State Obtain the distances $\mathcal{D}_o$ from each point to all bottom-layer cluster centers.
    			\EndIf
			\State Obtaining features $F_a$, $F_b$ and momentum feature $F^m_a$, $F^m_b$ respectively for current batch.
			\State Calculate distances from $F^m_a$ to $Q_a$ and from $F^m_b$ to $Q_b$
			\State Update $F^m_a$ and $F^m_b$ to $Q_a$ and $Q_b$, respectively, resulting in $Q_a^{update}$ and $Q_b^{update}$.
			\State Compute distances from $F_a$ to $Q_b^{update}$ and from $F_b$ to $Q_a^{update}$, apply angle normalization.
			\State Construct corresponding ground truth based on the insertion position of $F^m_a$ and $F^m_b$.
			\State Calculate InfoNCE loss and update model parameters $\theta$.
			\State Update $\mathcal{K}$, $\mathcal{D}_o$, $Q_a=Q_a^{update}$, $Q_b=Q_b^{update}$
		\EndFor
	\end{algorithmic}  
\end{algorithm}

\subsection{Implementation Details}
\label{sec_details}

We conduct our experiments with PyTorch framework on 8 NVIDIA V100 GPUs. Following~\cite{li2021align}, we extract textual features by the first 6 layers of a BERT~\cite{devlin2018bert} pre-trained text encoder, while visual features are obtained from an image encoder trained with CLIP-ViT~\cite{radford2021learning}. Then we compute the image-text contrastive loss $\mathcal{L}_{itc}$ on representations from the unimodal encoders. We replace the cosine distance in loss function with the proposed geodesic distance. Then we apply the last 6 layers of Bert model for multimodal fusion, where visual features interact with textual features through cross-attention. Additionally, we also employ two other pre-training tasks, i.e., masked language modeling (MLM) and image text matching (ITM), with loss functions $\mathcal{L}_{mlm}$ and $\mathcal{L}_{itm}$. The overall loss function is defined as $\mathcal{L} = \mathcal{L}_{itc} + \mathcal{L}_{mlm} + \mathcal{L}_{itm}$.

We empirically set the queue size (i.e., the sample pool size) as 65536 and update the hierarchical graph every 100 iterations. During the hierarchical graph construction, we utilized the K-Means~\cite{kmeans} for clustering with 256 cluster centers ($n_{cluster}$) and 5 clustering iterations. Each cluster center has 8 neighbors for constructing the graph. To balance efficiency and effectiveness, we build a two-layer graph structure, using cosine distance as the trivial distance metric, where the parameter $\delta$ in Def.\ref{def_simple_manifold} is set as $\sqrt{d}$ with feature dimension $d=256$. We truncate the final accumulated angle of the geodesic path at 4$\pi$. We conduct detailed experiments on parameter selection in Sec.~\ref{sec_hyper}. We employ the AdamW~\cite{adamw} optimizer with a weight decay of 0.02. For the learning rate schedule, we set the initial learning rate to $1e-5$, warm up to $1e-4$ after 1000 iterations, and then follow a cosine decay to the minimum learning rate of $1e-5$. We conduct pre-training for 30 epochs on COCO~\cite{lin2014microsoft}, VG~\cite{krishna2017visual}, SBU~\cite{ordonez2011im2text}, and CC3M~\cite{sharma2018conceptual} datasets, with data pre-processing following~\cite{li2021align}. We use Numba~\cite{numba} with CUDA operator to implement Floyd algorithm on each layer for computation acceleration.

More details about network structure, pre-training tasks can be found in the supplementary materials.

\begin{table*}[t]
\centering
\setlength\tabcolsep{10.6pt}
\begin{tabular}{l|c|cccccc}
\toprule
\multirow{2}{*}{Method} &\multirow{2}{*}{\#Images} & \multicolumn{2}{c}{VQA}         & \multicolumn{2}{c}{NLVR$^2$}       & \multicolumn{2}{c}{SNLI-VE}     \\
            &            & test-dev       & test-std       & dev            & test-P         & val            & test           \\ \hline
UNITER~\cite{chen2020uniter}    & 4M          & 72.70          & 72.91          & 77.18          & 77.85          & 78.59          & 78.28          \\
OSCAR~\cite{li2020oscar}   & 4M         & 73.16          & 73.44          & 78.07          & 78.36          & -              & -              \\
ViLT~\cite{kim2021vilt}     & 4M          & 70.94          & -              & 75.24          & 76.21          & -              & -              \\
VILLA~\cite{gan2020large}   & 4M         & 73.59          & 73.67          & 78.39          & 79.30          & 79.47          & 79.03          \\
ALBEF~\cite{li2021align}    & 4M     & 74.54   & 74.70  & 80.24  & 80.50   & 80.14     & 80.30          \\ 
TCL~\cite{yang2022vision}    & 4M   & 74.90   & 74.92  & 80.54   & 81.33    & 80.51    & 80.29          \\
MAFA~\cite{byun2024mafa}    & 4M    &  75.55   & 75.75  & 82.52  & 82.08    &  80.79   &  80.96     \\ 
\hline
\textbf{Geo-ALBEF}    & 4M          & 75.13 & 75.16 & 80.70 & 81.06 & 80.56 & 80.73 \\ 
\textbf{Geo-TCL}    & 4M          & 75.46 & 75.48 & 81.06 & 81.99 & 80.99 & 80.75 \\ 
\textbf{Geo-MAFA}   & 4M          & \textbf{76.04} & \textbf{76.19} & \textbf{83.12} & \textbf{82.95} & \textbf{81.42} & \textbf{81.63} \\\bottomrule
\end{tabular}
\caption{Performance on downstream visual-language tasks.}
\label{tab: vltasks}
\end{table*}

\subsection{Theoretical Analysis}
\label{sec_theoretical}

We conduct theoretical analysis of the hierarchical graph to verify \textbf{graph sparsity and connectivity}, which enables fast and robust geodesic distance calculation.

\begin{theorem}
\label{theo_number_comp}
(The number of connected components) For a layer of the hierarchical graph containing $N$ cluster centers, with each point having $\sigma$ neighbors, the number of connected components $\mathcal{E}$ in this layer satisfies:
\begin{equation}
\mathcal{E} <= \frac{N([1+\ln(1+\sigma)]}{1+\sigma} .
\end{equation}
\end{theorem}

\begin{theorem}
\label{theo_scale_comp}
(The scale of connected components) For a layer with $N$ cluster centers, the size of each connected component satisfies $b < |S_i| < a$, if and only if the number of edges in the graph satisfies:
\begin{equation}
\small
\label{eq_upper_scale}
b^2 \lfloor \frac{N}{b} \rfloor-N < E(N) \leq (a-1)^{\frac{1}{a}}N^{2-\frac{1}{a}}+(a-1)N.
\end{equation}
We can also deduce a similar upper bound with a more concise form,
\begin{equation}
\small
E(N) \leq (1-\frac{1}{a-1})\frac{N^2}{2},
\end{equation}
where $|\mathcal{S}_i|$ represent the scale of a certain connected components $\mathcal{S}_i$.
\end{theorem}

The above theorems indicate that each layer of the hierarchical graph may contain multiple connected and adequately large components, implying the rationality of computing geodesic distance in our multimodal learning setting. 
In our implementation, we set the distance between points that are not mutually reachable in the graph to infinity. This effectively filters out some overly simple negative samples. 
Detailed proof can be found in the supplementary materials.

\input{exp_abl.tex}

\section{Experiments}

\subsection{Downstream Tasks}
We evaluate our method on five downstream tasks, including image-text retrieval, visual question answering, visual entailment, natural language for visual reasoning, and visual grounding.

\noindent\textbf{Image-Text Retrieval} consists of two sub-tasks, image-to-text retrieval (TR) and text-to-image retrieval (IR). Following~\cite{li2021align}, we evaluate the pre-trained model on Flickr30K~\cite{plummer2015flickr30k} and COCO~\cite{lin2014microsoft} with both zero-shot and fine-tuning settings. For zero-shot retrieval on COCO, the pre-trained model is directly evaluated on the test data of COCO. For zero-shot retrieval on Flickr30K, we evaluate the model fine-tuned on COCO as ~\cite{li2021align}. Under fine-tuning setting, we fine-tune the pre-trained model on the training data of these datasets and evaluate on the corresponding validation and test sets.

\noindent\textbf{Visual Question Answering} (VQA) aims to predict the answer to a question based on the given image. This task can be considered as a generation task following~\cite{li2021align}. Specifically, we use a 6-layer decoder for answer generation from the 3192 candidates during inference.

\noindent\textbf{Visual Entailment} (SNLI-VE)~\cite{xie2019visual} predicts the relationship between the given image-text pair, which includes entailment, neutral, and contradictory. It can be treated as a three-class classification problem, and we add a multi-layer perceptron (MLP) to predict the class on the final \texttt{[CLS]} token. 

\noindent\textbf{Natural Language for Visual Reasoning} (NLVR$^2$) determines whether a natural language description matches a pair of images. We evaluate on the NLVR$^2$~\cite{suhr2018corpus} dataset following ~\cite{li2021align}.

\noindent\textbf{Visual Grounding} aims to localize the region in an image that is most relevant to the given text. We adopt the weakly-supervised setting similar to image-text retrieval and evaluate on RefCOCO+~\cite{yu2016modeling}. During inference, heatmaps generated by Grad-CAM~\cite{selvaraju2017grad} are used to rank the detected proposals.

\subsection{Image and Text Retrieval}

We evaluate the performance of our model on the COCO and Flickr30K  datasets under two settings, i.e., zero-shot and fine-tuning, as reported in Tab.\ref{tab: zero_re} and Tab.\ref{tab: ft_re}, respectively. We can observe that our method outperforms various arts, such as UNITER~\cite{chen2020uniter}, OSCAR~\cite{li2020oscar}, VILLA~\cite{gan2020large}, ViLT~\cite{kim2021vilt}, CLIP~\cite{radford2021learning}, ALBEF~\cite{li2021align}, VLMO~\cite{bao2022vlmo}, TCL~\cite{yang2022vision} and MAFA~\cite{byun2024mafa}, demonstrating the effectiveness of our method.

\subsection{VQA, NLVR2, VE and Grounding}
We further validate our method on various downstream tasks, including VQA, NLVR$^2$, VE and visual grounding, as shown in Tab.\ref{tab: vltasks}. Previous works like UNITER~\cite{chen2020uniter}, OSCAR~\cite{li2020oscar},  VILLA~\cite{gan2020large}, ALBEF~\cite{li2021align}, TCL~\cite{yang2022vision}, and MAFA~\cite{byun2024mafa} are compared.  Experimental results indicate satisfactory performance on these downstream tasks. We conduct visual grounding following~\cite{li2021align}, obtaining localization regions through Grad-CAM~\cite{selvaraju2017grad}. Fig.\ref{tab_fig: grounding}(right) presents our results on visual grounding, comparing with ARN~\cite{liu2019adaptive}, CCL~\cite{zhang2020counterfactual}, and ALBEF~\cite{li2021align}.

\subsection{Generalization}
\label{sec_ablation}

To further validate the effectiveness of geodesic distance, we replace distance metrics in other methods, such as CLIP~\cite{radford2021learning} and FLIP~\cite{li2023scaling}. To avoid time-consuming pre-training, we adapt their network to momentum queue form and fine-tune our pre-training dataset with the open-source model for an epoch, denoted as CLIP$_{FT}$ and FLIP$_{FT}$, where $Geo$ represents the use of geodesic distance, as shown in Tab.\ref{tab_generalize_multimodal}. We also verify the effectiveness of our method on self-supervised learning, such as MOCOv2~\cite{chen2020improved} and SwAV~\cite{caron2020unsupervised}, as shown in the Tab.\ref{tab_generalize_selfsupervision}. 
Under the default settings, geodesic distance achieves superior performance, showing its generalization capability.

\subsection{Other Analysis}
\label{sec_hyper}

We conduct detailed analyses of the hyper-parameters. For a fair comparison, we use pre-training models without geodesic distance and only fine-tuned using different hyper-parameters for the image-text retrieval, without affecting the validity of the experimental results. We analyze the number of hierarchical graph layers, the number of K-Means clusters, the number of neighbors for constructing the graph, and the frequency of hierarchical graph updates, as shown in Tab.~\ref{tab:ablations}. We conduct experiments on ALBEF~\cite{li2021align}, and similar results can be drawn from other models. It can be observed that our performance is not sensitive to those parameters. 
We also investigate the algorithm efficiency regarding training time and CUDA memory consumption, as shown in Tab.\ref{tab_algo_efficient}. We can observe that our method brings little extra time cost and a negligible increase in CUDA memory.

\section{Conclusion}
\label{sec_conclusion}

In this paper, we introduce geodesic distance for the first time to address the limitations of distance metrics in current multimodal learning frameworks.
Geodesic distance, as a more accurate measurement for large-scale datasets and complex data structures, enhances the effectiveness of negative sample mining in cross-modal contrastive learning. We illustrate the methodology of employing geodesic distance within the current multimodal learning paradigms, offering detailed adaptation strategies. To enhance the efficiency and applicability, we develop the hierarchical geodesic distance algorithm for computing geodesic distance along with corresponding update strategies. Extensive experiments on downstream tasks validate the effectiveness of our method.


\clearpage
\newpage

\noindent\textbf{Acknowledgement.} This work was supported by National Science Foundation of China (U20B2072). This work was also partly supported by SJTU Medical Engineering Cross Research Grant YG2021ZD18.


\input{X_suppl}

\end{document}

%% file: exp_abl.tex
\begin{figure*}[t]
\begin{minipage}{0.60\linewidth}
\centerline{\includegraphics[width=8.2cm,height=2.9cm]{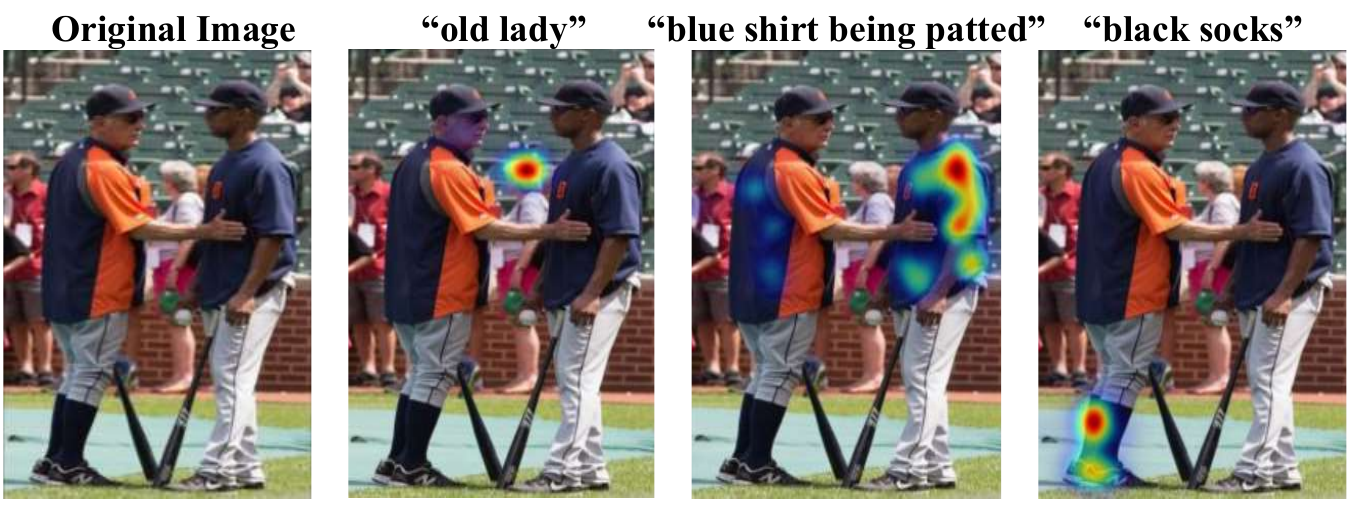}}
\end{minipage}
\begin{minipage}{0.35\linewidth}
\centering
\setlength\tabcolsep{5.0pt}
\begin{tabular}{l|ccc}
\toprule
Method  & Val & TestA & TestB \\ \hline
ARN~\cite{liu2019adaptive}    & 32.78  & 34.35 & 32.13 \\
CCL~\cite{zhang2020counterfactual}    & 34.29  & 36.91 & 33.56 \\
ALBEF~\cite{li2021align}  & 58.46  & 65.89 & 46.25 \\
\textbf{Geo-ALBEF}   & \textbf{59.82}  & \textbf{67.93} & \textbf{47.49} \\ \bottomrule
\end{tabular}

\end{minipage}
\vspace{-1mm}
\caption{Visual grounding visualization (left) and performance of weakly-supervised visual grounding on RefCOCO+ (right).}
\label{tab_fig: grounding}
\end{figure*}

\begin{figure*}[t] 
\begin{minipage}[!ht]{\textwidth}
\begin{minipage}[h]{0.31\textwidth}
\small
\makeatletter\def\@captype{table}\makeatother

\centerline
{
    \begin{tabular}{l|cc}
    \toprule
    Method & TR & IR \\ \hline
    CLIP$_{FT}$    & 58.5 & 37.8 \\
    Geo-CLIP$_{FT}$  & 59.6 & 38.7 \\
    FLIP$_{FT}$    & 60.1 & 44.2 \\
    Geo-FLIP$_{FT}$   & 61.7 & 45.1 \\ \bottomrule
    \end{tabular}
}
\caption{Generalization on other multimodal methods. We report R@1 accuracy of text retrieval (TR) and image retrieval (IR) on COCO with zero-shot setting.}
\label{tab_generalize_multimodal}
\end{minipage}
\hspace{1.2mm}
\begin{minipage}[h]{0.31\textwidth}
\small
\makeatletter\def\@captype{table}\makeatother

\centerline
{   
    \setlength\tabcolsep{2.3pt}
    \begin{tabular}{l|cc}
    \toprule
    Method & 100epoch & 200epoch \\ \hline
    MOCOv2    & 67.4 & 69.9 \\
    Geo-MOCOv2  & 67.8 & 70.6 \\
    SwAV    & 66.5 & 69.1 \\
    Geo-SwAV   & 67.1 & 69.9 \\ \bottomrule
    \end{tabular}
}
\caption{Generalization on other self-supervised methods. We report the Top1 accuracy of linear classification on ImageNet dataset. (ResNet50 crop 224$\times$224).}
\label{tab_generalize_selfsupervision}
\end{minipage}
\hspace{1.2mm}
\begin{minipage}[h]{0.28\textwidth}
\small
\makeatletter\def\@captype{table}\makeatother

\centerline
{
    \begin{tabular}{l|cc}
    \toprule
    Method & PCMA & TTI \\ \hline
    ALBEF    & 25.3G & 255s \\
    Geo-ALBEF  & 25.9G & 268s \\
    CLIP$_{FT}$    & 24.3G & 213s \\
    Geo-CLIP$_{FT}$   & 24.7G & 224s \\ \bottomrule
    \end{tabular}
}
\caption{Efficiency analysis. PCMA represents the peak CUDA memory allocations, and TTI denotes the total time cost of 100 training iterations.}
\label{tab_algo_efficient}
\end{minipage}
\end{minipage}
\end{figure*}

\begin{table*}[!htb]
\small
\vspace{-1mm}
\label{tab:ablation}
\subfloat[
\textbf{Graph layers}.
\label{tab:Graph_layers}
]{
\centering
\begin{minipage}{0.23\linewidth}{\begin{center}
\setlength\tabcolsep{2.8pt}
\begin{tabular}{x{20}x{20}x{20}x{20}}
layers & TR & IR & TTI \\
\toprule
1 & 75.1 & 58.5 & 282s \\
2 & \textbf{76.2} & \textbf{59.2} & 268s \\
3 & \textbf{76.2} & 59.1 & 270s \\
\bottomrule
\end{tabular}
\end{center}}\end{minipage}
}
\hspace{0.8em}
\subfloat[
\textbf{Kmeans clusters}. 
\label{tab:Kmeans_clusters}
]{
\begin{minipage}{0.21\linewidth}{\begin{center}
\setlength\tabcolsep{3.2pt}
\begin{tabular}{x{24}x{24}x{24}}
clusters & TR & IR \\
\toprule
128 & 74.7 & 58.1 \\
192 & 75.3 & 58.8 \\
256 & \textbf{76.2} & \textbf{59.2} \\
384 & 75.8 & 59.0 \\
\bottomrule
\end{tabular}
\end{center}}\end{minipage}
}
\hspace{0.8em}
\subfloat[
\textbf{Point neighbors}. 
\label{tab:Point_neighbors}
]{
\centering
\begin{minipage}{0.21\linewidth}{\begin{center}
\setlength\tabcolsep{3.2pt}
\begin{tabular}{x{18}x{24}x{24}}
nbrs & TR & IR \\
\toprule
6 & 75.9 & 58.6 \\
8 & \textbf{76.2} & \textbf{59.2} \\
10 & 76.1 & 59.0 \\
12 & \textbf{76.2} & 59.1 \\
\bottomrule
\end{tabular}
\end{center}}\end{minipage}
}
\hspace{0.8em}
\subfloat[
\textbf{Graph updates}. 
\label{tab:Graph_updates}
]{
\centering
\begin{minipage}{0.21\linewidth}{\begin{center}
\setlength\tabcolsep{3.3pt}
\begin{tabular}{x{24}x{24}x{24}}
iters & TR & IR \\
\toprule
50 & 75.9 & 59.1 \\
75 & 76.0 & 59.1 \\
100 & \textbf{76.2} & \textbf{59.2} \\
125 & 76.0 & 58.9 \\
\bottomrule
\end{tabular}
\end{center}}\end{minipage}
}
\centering
\caption{Hyper-parameter analysis. We present the R@1 accuracy of text retrieval (TR) and image retrieval (IR) on the COCO dataset.}
\label{tab:ablations} 
\vspace{-2mm}
\end{table*}

%% file: X_suppl.tex
\clearpage
\setcounter{page}{1}
\maketitlesupplementary

\section{Network Structure}
Our model structure mainly follows~\cite{li2021align}, \cite{yang2022vision} and \cite{byun2024mafa}. Taking \cite{li2021align} as an example, the network includes an image encoder, a text encoder, and a multimodal fusion encoder. The image encoder is a 12-layer transformer with a VIT structure, with initialized weights derived from pre-training on the ImageNet-1k~\cite{deng2009imagenet} dataset. The text encoder and fusion encoder use pre-trained BERT-base~\cite{devlin2018bert} networks, which consist of 12 layers of transformers, with each encoder using 6 layers. To obtain more robust training with noisy web datasets, we maintain the momentum version for each encoder, i.e., $\theta_t' = \alpha \theta_{t-1}' + (1-\alpha)\theta_t$, where $\theta_t$ and $\theta_t'$ are the parameters of the main model $E_I$ and the momentum modal $E_I'$, and $\alpha$ is the momentum parameter ranging between $[0,1]$. Given a pair of images and text $(I, T)$, following \cite{he2020momentum}, we first perform two data augmentations on the image to obtain two different views of the image. Then, we encode the two augmented images with the original model and momentum model separately as positive pairs. The image tokens are obtained after the image patches are linearly mapped and position encoded. To capture global features, we also concatenate CLS tokens before visual tokens. We encode $I$ and $I'$ with $E_I$ and $E_I'$ respectively to obtain image embeddings, $V=\{V_{cls}, V_1, V_2, ..., V_m\}$ and $V'=\{V_{cls}', V_1', V_2', ..., V_m'\}$. For the text, we tokenize and embed text following BERT\cite{devlin2018bert}, and we can similarly obtain text embeddings $W=\{W_{cls}, W_1, W_2, ..., W_n\}$ and $W'=\{W_{cls}', W_1', W_2', ..., W_n'\}$. Subsequently, we concatenate visual and textual embeddings and feed them together into the fusion encoder for feature fusion, thereby learning the joint modal representations.

\section{Pre-training Datasets}

The details of the pre-training datasets about image-text pairs are shown in below Tab.\ref{fig: predata_stat}. 
\begin{table}[h!]
\centering
\begin{tabular}{l|cccc}
\hline
        & COCO & VG   & SBU  & CC3M    \\ \hline
\#image & 113K & 100K & 860K & 2.95M \\
\#text  & 567K & 769K & 860K & 2.95M  \\ \hline
\end{tabular}
\caption{Statistics of the pre-training datasets.}
\label{fig: predata_stat}
\end{table}

\section{Pre-training Tasks}

Multimodal learning requires elaborate pre-training tasks, and commonly used pre-training tasks include Masked Language Modeling(MLM)~\cite{devlin2018bert}, Image-Text Matching (ITM)~\cite{li2021align}, Image-Text Contrastive (ITC)~\cite{radford2021learning}, Word Patch Alignment (WPA)~\cite{chen2020uniter}, and so on. We leverage MLM, ITM, and ITC for multimodal pre-training following~\cite{chen2020uniter, kim2021vilt, li2021align}.

\section{Extra Experiments}

We also compare our method with the Oblique manifold ($\mathcal{OM}$)~\cite{andruchow1999geometry}. We conduct the experiments on the image-text retrieval task with the COCO dataset and fine-tune setting. We display the R@1 accuracy for text retrieval (TR) and image retrieval (IR), as shown in Tab.\ref{tab_comp_om}.
\begin{table}[]
    \centering
    \setlength\tabcolsep{10pt}
    \begin{tabular}{l|cc}
    \toprule
    Method & TR & IR \\ \hline
    ALBEF    & 73.1 & 56.8 \\
    ALBEF+$\mathcal{OM}$    & 73.3 & 56.4 \\
    ALBEF+Geo  & \textbf{76.2} & \textbf{59.2} \\ \hline
    MAFA    & 78.0  & 61.2 \\
    MAFA+$\mathcal{OM}$    & 77.8 & 61.1 \\
    MAFA+Geo  & \textbf{79.3} & \textbf{62.5} \\
     \bottomrule
    \end{tabular}
    \caption{Comparison with Oblique manifold.}
    \label{tab_comp_om}
\end{table}

\section{Proofs}

\subsection{Proof for Theorem 1}

For the graph where the cluster centers represent the vertices of the graph and the adjacent relationship between these cluster centers represents the edges between vertices, we can know this graph possesses $N$ vertices with minimum degree $\kappa$ (cluster neighbors).

We define all the vertices set as $V$, and then we construct a random subset $X$ of $V$ ($X \subset V$). Each sample in $X$ is taken from $V$ with a probability of $p$. Then the expectation scale of $X$ is,
\begin{equation}
\mathbb{E}(|X|) = Np
\end{equation}
We regard the subset $X$ as the candidate for connected components $\mathcal{S}$. We can thus define the random set $Y_X$, which represents the samples in $V-X$ that do not have an adjacent sample in $X$, that is, for sample $v \in Y_X$, we can not find a sample $x\in X$ that $v$ is subordinate to $x$. This can also be interpreted as for $v\in Y_X$, any adjacent samples of $v$ not in $X$, so
\begin{equation}
\begin{split}
P(v\in Y_X) &= P(\text{v and its adjacent samples not in X}) \\
&=(1-p)^{1+d(v)} \\
&\leq (1-p)^{1+\kappa}
\end{split}
\end{equation}
Then we can obtain,
\begin{equation}
\mathbb{E}(|Y_X|) \leq N(1-p)^{1+\kappa}
\end{equation}

It is apparent that $X\cup Y_X$ can be served as a connected component, and the number of connected components can be represented as,
\begin{equation}
\begin{split}
\mathbb{E}(|X\cup Y_X|) &\leq \mathbb{E}(|X|+|Y_X|) \leq \mathbb{E}(|X|) + \mathbb{E}(|Y_X|)    \\
&= Np + N(1-p)^{1+\kappa} \leq Np + Ne^{-p(1+\kappa)}
\end{split}
\end{equation}

Since we want to find the minimal number of connected components, which means we want to find the minimal value of $Np + Ne^{-p(1+\kappa)}$. We can then obtain that when $p=\frac{\ln (\kappa +1)}{\kappa +1}$, the expectation get the minimum value,
\begin{equation}
\frac{N[1+\ln (\kappa +1)]}{\kappa +1}
\end{equation}
Therefore we can get
\begin{equation}
|\mathcal{S}| \leq \frac{N[1+\ln (\kappa +1)]}{\kappa +1},\quad \text{where} \quad \kappa = \mathcal{F}(\xi).
\end{equation}

\subsection{Proof for Theorem 2}

The lower bound is obvious. Here we only prove the upper bound.

We can model the adjacency relationship between cluster centers as a bipartite graph with $N$ points on both sides. The problem can be transformed into a $0-1$ matrix of $\mathcal{M} \in \mathbb{R}^{N\times N}$, where there is no all 1 sub-matrix of $\mathcal{M}_0 \in \mathbb{R}^{A\times A}$, and at this time, how many element 1 can there be in the matrix at most.

We count the following structures. We define that the left and right point sets of the bipartite graph are $V_1$ and $V_2$, and then assume that the structure $p$ is selecting a point $u$ from $V_1$ with $a$ adjacent samples in $V_2$. Let's start with point $u$ in $V_1$, and the selection methods of $a$ samples in $V_2$ is $C_{d(u)}^a$, then the total selections are $\sum_{u\in V_1}C_{d(u)}^a = |S|$. We can also start with $a$ samples in $V_2$.  And once we determine $a$ point in $V_2$, there are at most $a-1$ $u$ in $V_1$, otherwise there will be an all 1 sub-matrix of $\mathcal{M}_0 \in \mathbb{R}^{A\times A}$. We can thus obtain,
\begin{equation}
\sum_{u\in V_1}C_{d(u)}^a \leq C_N^a(a-1)
\end{equation}
Following Jensen Inequality~\cite{mcshane1937jensen}, $f(x)=C_x^a$ is a convex function, then,
\begin{equation}
\sum_{u\in V_1}\frac{1}{N}C_{d(u)}^a \geq C_{\frac{1}{N}\sum_{u\in V_1}d(u)}^a = C_{\frac{|E|}{a}}^a
\end{equation}
So,
\begin{equation}
\begin{split}
N C_{\frac{|E|}{a}}^a &\leq \sum_{u\in V_1}C_{d(u)}^a \leq C_N^a(a-1)  \\
&= \frac{N(N-1)...(N-a+1)}{a!}(a-1)
\end{split}
\end{equation}
Retraction is conducted at both sides and then,
\begin{equation}
N\frac{(\frac{|E|}{N}-a+1)^a}{a!} < N C_{\frac{|E|}{a}}^a < \frac{N^a}{a!}(a-1)
\end{equation}
After simplification, we can obtain,
\begin{equation}
E(N) \leq (a-1)^{\frac{1}{a}}N^{2-\frac{1}{a}}+(a-1)N
\end{equation}

The proof of another upper bound is presented as follows,

Let the number of cluster centers be $N$ and for every point $x_i$, the number of neighboring points is $d(x_i)$. Suppose a initial set $C_{\pi}=\varnothing$, for all points, we introduce a random permutation $\mathcal{O}:x_1,x_2,x_3,...,x_n$. For a certain permutation, if all points in front of $x_i$ are $x_i$'s neighboring points, we put $x_i$ into $C_{\pi}$. Finally, all point pairs in $C_{\pi}$ are neighboring points.

The probability of a certain point in $C_{\pi}$ is $p = \frac{1}{N-d(x_i)}$, then the mathematical expectation of the size of $C$ is,
\begin{equation}
| C_{\pi} | = \sum_{x_i} \frac{1}{N-d(x_i)}
\end{equation}
Suppose the size of maximal cluster is $\omega (D)$, we apply Pigeonhole Principle~\cite{trybulec1990pigeon} and get:
\begin{equation}
\omega(\mathcal{D}) \geq \sum_{x_i} \frac{1}{N-d(x_i)}
\end{equation}

What we need to satisfy is,
\begin{equation}
a \geq \omega (G) \geq \sum_{v_i} \frac{1}{N-d(v_i)}
\end{equation}
According to Cauchy Inequality~\cite{mitrinovic1970analytic},
\begin{equation}
a\sum_{v_i}(N-d(v_i)) \geq \sum_{v_i} \frac{1}{N-d(v_i)}\sum_{v_i}(N-d(v_i)) \geq N^2
\end{equation}
So,
\begin{equation}
a(N^2-2|E|) \geq N^2
\end{equation}
We can thus obtain,
\begin{equation}
|E| \leq \frac{N^2}{2}(1-\frac{1}{a-1})
\end{equation}